\documentclass[sigconf]{acmart}

\usepackage{booktabs} 
\usepackage{todonotes}
\setcopyright{rightsretained}
\usepackage{colortbl}
\usepackage{rotating}

\usepackage{array}
\newcolumntype{L}[1]{>{\raggedright\let\newline\\\arraybackslash\hspace{0pt}}m{#1}}
\newcolumntype{C}[1]{>{\centering\let\newline\\\arraybackslash\hspace{0pt}}m{#1}}
\newcolumntype{R}[1]{>{\raggedleft\let\newline\\\arraybackslash\hspace{0pt}}m{#1}}

\def\tw{1.8cm}

\def\sp{\,\,\,\,\,\,\,\,\,\,\,\,\,\,\,\,\,}
\def\spb{\,\,\,\,\,\,\,\,\,\,\,\,\,\,\,}
\def\spc{\,\,\,\,\,\,\,\,\,}
\def\translation{\textcolor{red}{$\blacksquare$}}
\def\scale{\textcolor{blue}{$\blacksquare$}}
\def\rotation{\textcolor{green}{$\blacksquare$}}
\def\perspective{\textcolor{cyan}{$\blacksquare$}}
\def\matchPart{\textcolor{brown}{$\blacksquare$}}
\def\multiInst{\textcolor{yellow}{$\blacksquare$}}
\def\multiSpli{\textcolor{orange}{$\blacksquare$}}
\def\simiBg{\textcolor{black}{$\blacksquare$}}
\def\simiFg{\textcolor{gray}{$\blacksquare$}}

\newcommand{\notsure}[1]{\textcolor{black}{#1}}
\newcommand*\rot{\rotatebox[origin=c]{90}}

\acmDOI{10.475/123_4}

\acmISBN{123-4567-24-567/08/06}

\acmConference[Arxiv Preprint]{}{}{} 
\acmYear{2017}
\copyrightyear{2017}

\graphicspath{{./images/}}

\begin{document}
\title[Deep Matching and Validation Network]{Deep Matching and Validation Network}
\subtitle{An End-to-End Solution to Constrained Image Splicing Localization and Detection}

\author{Yue Wu}
\authornote{Corresponding author}
\orcid{1234-5678-9012}
\affiliation{%
  \institution{Information Sciences Institute}
  \streetaddress{4676 Admiralty Way}
  \city{Marina Del Rey} 
  \state{CA} 
  \postcode{90292}
}
\email{yue_wu@isi.edu}

\author{Wael AbdAlmageed}
\affiliation{%
  \institution{Information Sciences Institute}
  \streetaddress{4676 Admiralty Way}
  \city{Marina Del Rey} 
  \state{CA} 
  \postcode{90292}
}
\email{wamageed@isi.edu}
\author{Prem Natarajan}
\orcid{1234-5678-9012}
\affiliation{%
  \institution{Information Sciences Institute}
  \streetaddress{4676 Admiralty Way}
  \city{Marina Del Rey} 
  \state{CA} 
  \postcode{90292}
}
\email{pnataraj@isi.edu}

\renewcommand{\shortauthors}{R. Wu et al.}

\begin{abstract}
Image splicing is a very common image manipulation technique  that is sometimes used for malicious purposes. A splicing detection and localization algorithm usually takes an input image and produces a binary decision indicating whether the input image has been manipulated, and also a segmentation mask that corresponds to the spliced region. Most existing splicing detection and localization pipelines suffer from two main shortcomings: 1) they use handcrafted features that are not robust against subsequent processing (e.g., compression), and 2) each stage of the pipeline is usually optimized independently. In this paper we extend the formulation of the underlying splicing problem to consider two input images, a query image and a potential donor image. Here 
the task is to estimate the probability that the donor image has been used to splice the query image, and obtain the splicing masks for both the query and donor images. We introduce a novel deep convolutional neural network architecture, called Deep Matching and Validation Network (DMVN), which simultaneously localizes and detects image splicing. The proposed approach does not depend on handcrafted features and uses raw input images to create deep learned representations. Furthermore, the DMVN is end-to-end optimized to produce the probability estimates and the segmentation masks. Our extensive experiments demonstrate that this approach outperforms state-of-the-art splicing detection methods by a large margin in terms of both AUC score and speed. 
\end{abstract}

%


\def\eg{\emph{e.g.\,}}
\def\ie{\emph{i.e.\,}}
\def\etal{\emph{et al.\,}}
\def\wrt{\emph{w.r.t.\,}}
\newcommand{\im}[1]{\mathsf{#1}}

\maketitle

\section{Motivation}\label{sec.motivation}
The ubiquity of digital cameras and the rapid growth of social networks have caused a proliferation of image and video content. Image forgery is becoming a rampant problem, as a direct consequence of digital content proliferation. Literally, the common idiom \emph{seeing is believing} does not hold true anymore, especially since in recent years sophisticated image editing tools, such as Adobe PhotoShop\texttrademark\ and GIMP have been pushing the limits of image composition in order to produce more natural and aesthetic images. These tools make it much easier to alter an image maliciously for a non-professional user. Meanwhile, detecting and localizing image forgeries, at a large scale, is becoming increasingly more difficult for new processionals \cite{zampoglou2016web}, forensic experts, and legal prosecutors. These new challenges necessitate developing novel and scalable image forensics technologies. 

\begin{figure}[!h]
\scriptsize
\centering
\begin{tabular}{@{}c@{}c@{}c@{}}
\includegraphics[width=.3\linewidth]{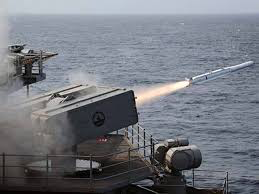}(a)&
\includegraphics[width=.3\linewidth]{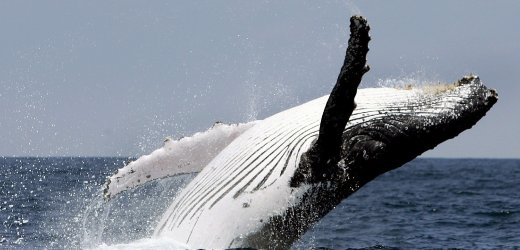}(b)&
\includegraphics[width=.3\linewidth]{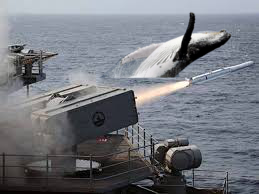}(c)\\
\includegraphics[width=.3\linewidth]{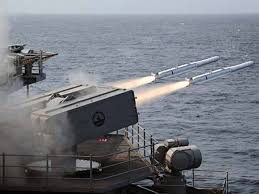}(d)&
\includegraphics[width=.3\linewidth]{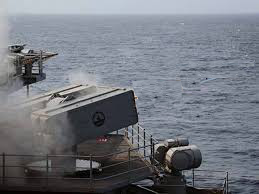}(e)&
\includegraphics[width=.3\linewidth]{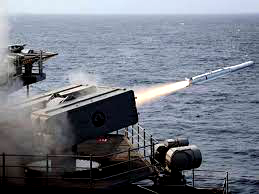}(f)\\
\end{tabular}
\caption{Image manipulation types. (a) and (b) original \textit{missile} and \textit{whale} images; (c) spliced image by compositing \textit{whale} into \textit{missile}; and (d)-(f) resulting images on \textit{missile} after \textit{copy-move}, \textit{erasing}, and \textit{retouching}, respectively.}\label{fig.manipulation}
\end{figure}

Although \emph{image manipulation} is sometimes used to indicate any kind of technique that can be used to modify an image, it often means major manipulations from an image forensics perspective \cite{asghar2016copy,qureshi2015bibliography}, such as \emph{splicing}, \emph{copy-move}, \emph{erasing}, or \emph{retouching}. Fig.~\ref{fig.manipulation} illustrates these common manipulations, where \textit{splicing} denotes copying one or more source image regions and pasting them onto a destination image, while the other three types can be done using a single source image. Thus, \emph{splicing} is considered to be more complicated since it involves external images. 

Traditionally, copy-move and image splicing forgery detection are often thought of as two close problems \cite{amerini2011sift,asghar2016copy} that can be solved within a general forgery detection framework (GFDF) \cite{cozzolino2015efficient,christlein2012evaluation,cozzolino2015splicebuster}, that is: 1) \textbf{representation}, in which the characteristics of the underlying image (pixel by pixel or holistically) are extracted as feature vectors; 2) \textbf{matching}, in which corresponding regions are determined from the feature representation; and 3) \textbf{post-processing}, in which nearest-neighbor detection is linked and filtered to reduce false alarms and improve detection rates. It is worth noting that many copy-move and image splicing detection algorithms rely on strong or specific image hypothesis, e.g., photo-response non-uniformity noise~\cite{chen2008determining,pan2011exposing}, camera characteristics~\cite{hsu2006detecting}, color filter array~\cite{swaminathan2008digital}, JPEG compression~\cite{chen2011detecting,bianchi2012image,amerini2014splicing,Liu:2014:IAC:2699158.2560365}, edge sharpness~\cite{qu2009detect,fang2010image} and local features~\cite{zhu2016copy,li2017image,li2017imageQDCT}. Comprehensive reviews of these approaches can be found in \cite{farid2009image, asghar2016copy,warif2016copy,birajdar2013digital}. The main assumption of the approaches, in order to achieve high detection rates, is that one or more of these clues must be present in a spliced image. However, this assumption is not always valid since splicing manipulations are usually followed by transformations (e.g., compression, resampling or geometric transformations) that may hide traces of the manipulation \cite{asghar2016copy,barni2012universal}. 

\begin{figure}[!h]
\scriptsize
\centering
\includegraphics[width=1\linewidth]{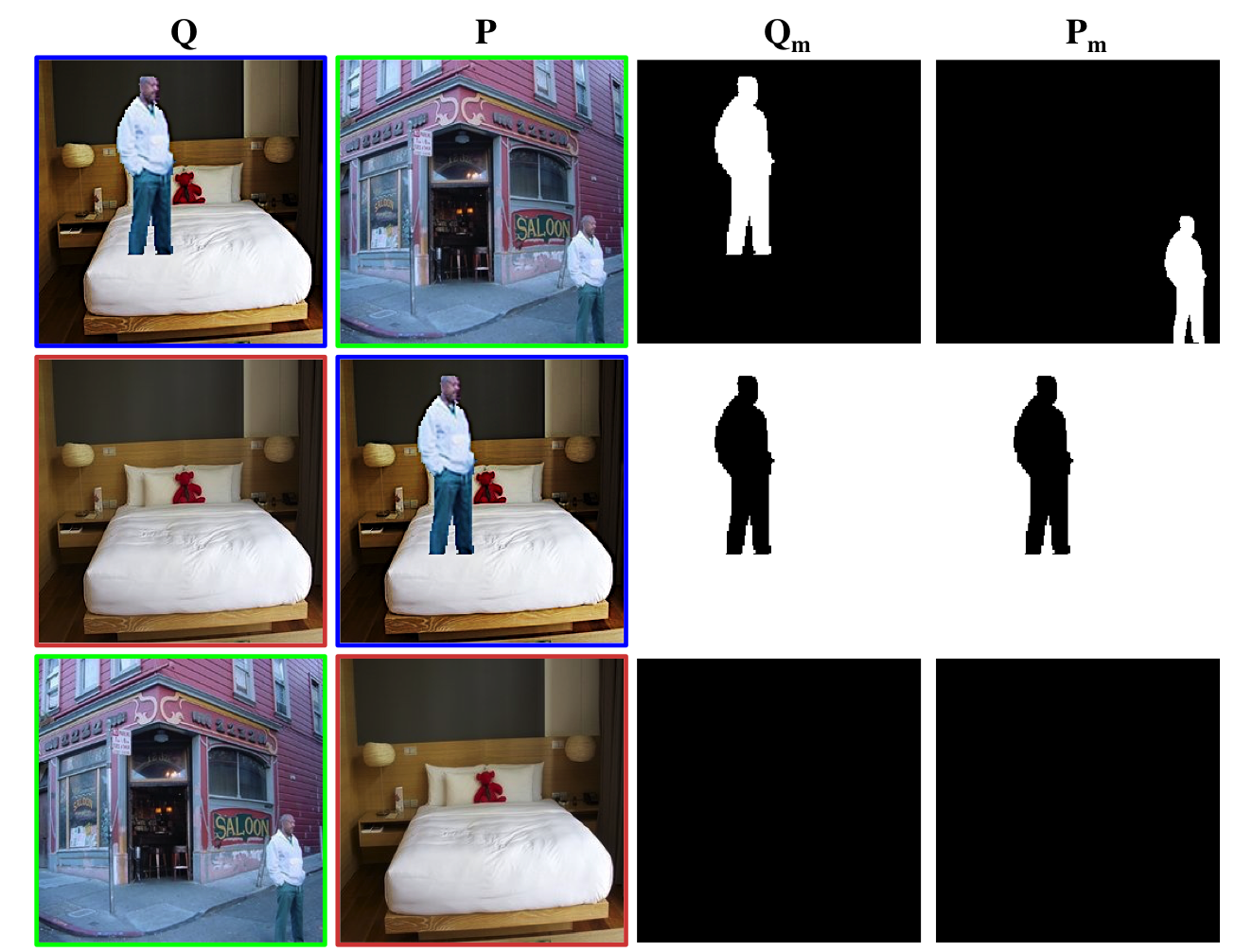}
\caption{Constrained image splicing detection problem, where true spliced pixels are labeled as white. From top to bottom, sample detection labels are 1, 1, and 0, respectively.}\label{fig.samples}
\end{figure}

\notsure{
In the recent Nimble 2017 Challenge from National Institute of Standards and Technology,\footnote{\url{https://www.nist.gov/itl/iad/mig/nimble-challenge-2017-evaluation}} the image splicing problem has been reformulated as: given a query image $\im{Q}$ and a potential donor image $\im{P}$, the goal is to solve not only the detection problem, i.e., whether or not $\im{Q}$ contains spliced regions from $\im{P}$, but also the localization problem, i.e., segmenting the spliced region(s) in both the donor and the spliced images. Since this new problem formulation constrains image splicing detection to a pair of images, we refer to it as the constrained image splicing detection (CISD) problem. Fig.~\ref{fig.samples} shows three input samples along with their ground truth splicing masks and detection labels of CISD. This CISD problem can be viewed as a new formulation of the classic copy-move detection problem in the sense that it also looks for a potential region that is copy-move from image $\im{P}$ to image $\im{Q}$. Finally, this new CISD problem also plays an important role in producing an image phylogeny graph \cite{dias2012image,dias2013large,de2016multiple} for a query image given a big dataset, especially in explaining how two images in neighboring nodes are associated.}

It is worth noting that the two-input nature of the new CISD problem makes many existing image hypothesis used in classic copy-move and splicing detection no longer applicable. For example, \cite{amerini2014splicing} proposed an image splicing detection algorithm by differentiating single and double JPEG compression, but it is not useful for the CISD problem because 1) we can neither guarantee that the two inputs will be in JPEG format, nor that the inputs are compressed with the same quantization table at the same level of quality; and 2) even though both are of JPEG format and one region in $\im{P}$ is detected as doubly compressed, the CISD question ``whether this region is originally from $\im{Q}$'' is still unanswered. As a result, the new CISD definition urges the use of more robust assumptions and features. 

\notsure{
Fortunately, visual clues, the visual correspondences between splicing regions~\cite{wang2009effective,zhao2012optimal,zhu2016copy,li2017imageQDCT,ardizzone2015copy,li2015segmentation,pun2015image}, are still useful in the new CISD problem because they are probably the weakest assumptions that we can make for that problem. This implies that we need two things: 1) representations for visual clues, and 2) rules to determine which two representations match. As one can see, these are exactly the first two steps (``representation'' and ``matching'') in the GFDF, while the last step (``post-processing'') in the GFDF is really to take advantage of the consistency within a set of true matchings to reduce false alarms. Though it seems that classic copy-move and splicing detection algorithms ~\cite{zhu2016copy,li2017image,li2017imageQDCT,wang2009effective,zhao2012optimal} relying on visual features can be easily modified for the new CISD problem, we note that the two major drawbacks of these existing algorithms are: 1) handcrafted features are less robust against image transformations (e.g., compression, noise addition and geometric transformations) and are surely not optimal for the CISD problem; and 2) tuning each of three stages in GFDF on its own only optimizes performance disjointly instead of jointly. }

\begin{figure*}[tbph]
\scriptsize
\centering
(a) The proposed DMVN pipeline. \\
\includegraphics[trim=0cm 0cm 0cm .3cm, clip, width=.85\linewidth]{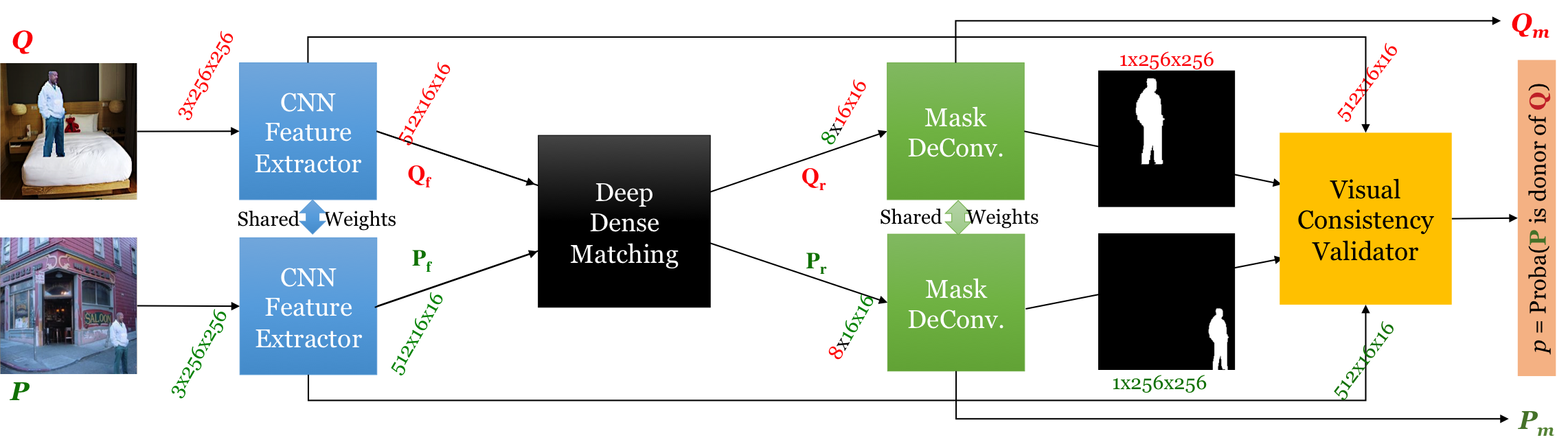}\\
(b) Modules of deep dense matching, mask deconvolution, and visual consistency validator. 
\includegraphics[width=.9\linewidth]{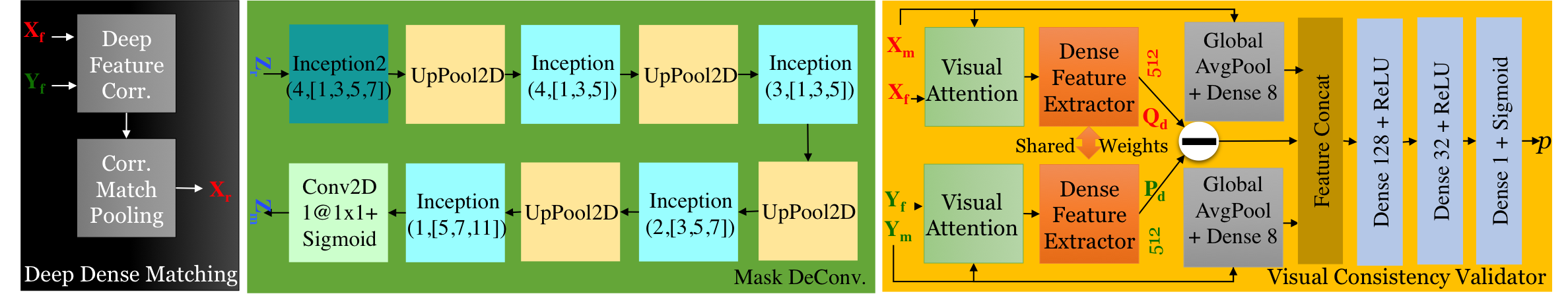}\\
\caption{Deep matching and validation network for the constrained image splicing detection and localization.}\label{fig.pipeline}
\end{figure*}
In this paper we conceptually follow the GFDF and propose Deep Matching and Verification Network (DMVN)---a novel deep learning-based splicing detection and localization method that is 1) unlike previous GFDF approaches, end-to-end optimized, 2) does not depend on extracting handcrafted, unrobust feature representations, 3) uses fully learnable parameters to determine matching or not, and 4) mimics the human validation process to see whether the found visual evidence is enough to determine a detection. 
\notsure{Furthermore, the proposed method is also distinct from recent deep learning based forgery detection practices~\cite{zhangDNN,rao2016deep,cozzolino2016single,cozzolino2017recasting} in the sense that our approach 1) is a full end-to-end deep learning solution instead of a deep learning module only for feature extraction~\cite{cozzolino2017recasting,cozzolino2016single}, 2) performs both localization and detection tasks instead of one or the other~\cite{zhangDNN,cozzolino2016single,rao2016deep}, and 3) invents unique deep learning modules (\emph{Deep Dense Matching} and \emph{Visual Consistency Validator}) for performing visual matching and validation (see Fig.\ref{fig.pipeline}-(a)). The remainder of this paper is organized as follows. Section \ref{sec.ddmn} describes the proposed DMVN and discusses the training procedure and settings. Experimental results and comparisons against state-of-the-art methods are presented in Section \ref{sec.experiments}. In Section \ref{sec.conclusions}, we conclude the paper and provide directions for future research.}


\section{Deep Matching Validation Network}\label{sec.ddmn}
\subsection{Architecture Overview}
As previously mentioned, the CISD problem is formulated as follows: given a query image $\im{Q}$ and a potential donor image $\im{P}$, we need to determine whether the query image is indeed spliced and then segment the spliced region in the query image and its corresponding region in the donor image.

As shown in Fig. ~\ref{fig.pipeline}, the overall network is designed such that block-wise learned representations of the input images are extracted using a convolutional network network -- \emph{CNN Feature Extractor} (e.g., AlexNet, ResNet or VGG) and fed into the proposed \emph{Deep Dense Matching} module, which performs (as the name implies) dense matching between the two input images. In order to segment the splicing masks in the two images, we use an inception-based \emph{Mask Deconvolution} module ~\cite{inception}. Further, the predicted masks are fed into a \emph{Visual Consistency Validator} module that forces the model to focus on the segmented areas in both images. Finally, a Siamese-like module is used to extract splicing-specific dense representations of the segmented regions in the donor and query images, and it produces a probability value indicating the likelihood that the donor image was used to splice the query image. We describe the details of each of these stages in the following. 

\subsection{Splicing Localization}
Although other CNN models (e.g., ResNet \cite{He2015}) can be also applied, we use the first four convolutional blocks of the VGG16 model~\cite{vgg16} just for the sake of simplicity. Consequently, the two network inputs $\im{Q}$ and $\im{P}$ (of shape $3\times 256\times 256$) are transformed into deep tensor representations $\im{Q}_f$ and $\im{P}_f$ (of shape $512\times 16\times16$.) It is well known that CNN features like $\im{Q}_f$ and $\im{P}_f$ have already exhibited certain level of invariance to luminance, scale, and rotation.

\begin{figure*}[tbph]
\scriptsize
\centering
\begin{tabular}{c@{}c}
\includegraphics[width=.76\linewidth]{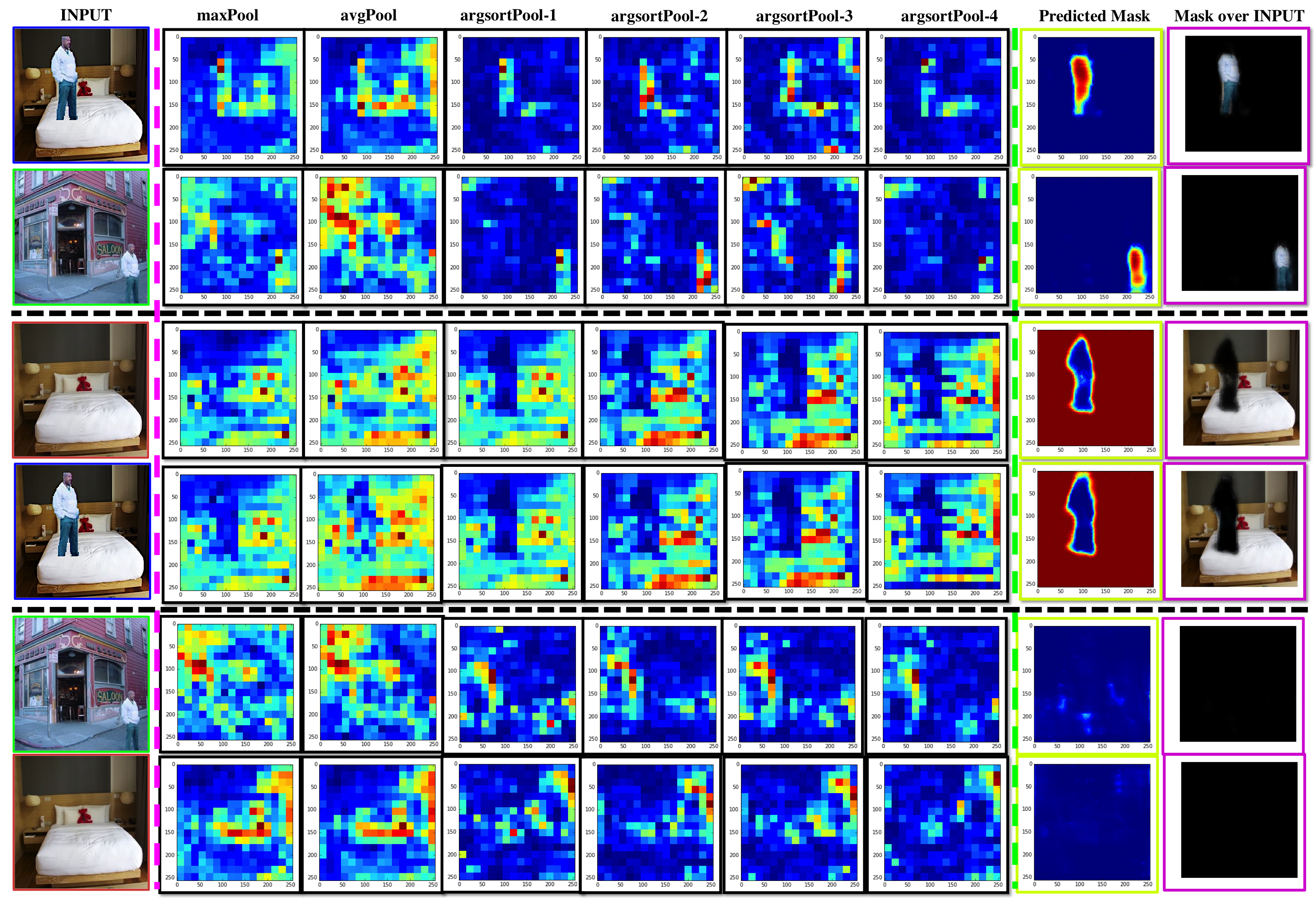}&
\includegraphics[width=.1729\linewidth]{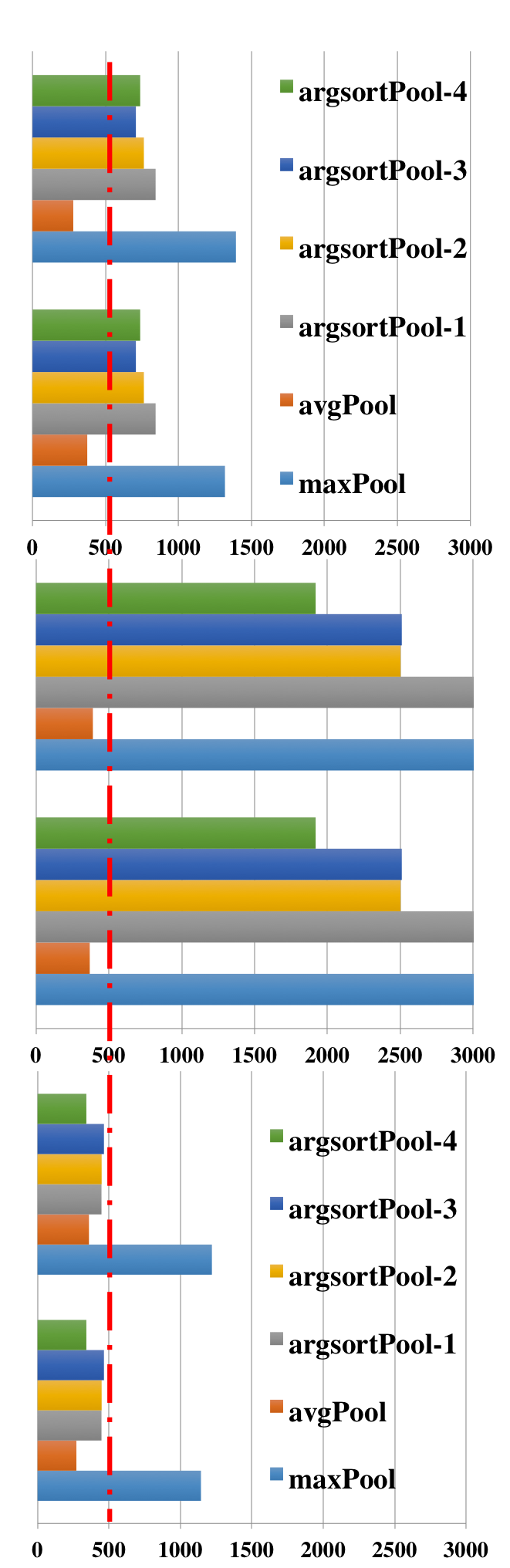}\\
\end{tabular}
\caption{Visualization of selected DMVN layers and the statistics of the 5th maximum responses from correlation matching pooling layers. Note: 1) all layers are shown by linearly rescaling to [0,1], and visualized \wrt the \texttt{Jet} color map, where more red means higher response; 2) data ranges of pooled matching response maps could be very different, but predicted masks share the same range [0,1]; and 3) rough maximum values of matching responses can be seen in the statistics on the far-right.
}\label{fig.visPooling}
\end{figure*}
The purpose of \emph{Deep Dense Matching} is to find possible matching regions between representations $\im{Q}_f$ and $\im{P}_f$. As shown in Fig.~\ref{fig.pipeline}(b), this is achieved through two steps, namely \emph{Deep Feature Correlation} and \emph{Correspondence Match Pooling}. In \emph{Deep Feature Correlation}, we exhaustively compute matching response using cross-correlation over all possible translations, as shown in Eq. \eqref{eq.corr}
\begin{equation}\label{eq.corr}
\textrm{corr}(\im{P}_f,\im{Q}_f)[x,y,i,j]=\textrm{trans}(\im{P}_f,x,y)[:,i,j] \cdot \im{Q}_f[:,i,j]
\end{equation}
where $\cdot$ is the dot product operator, and $\textrm{trans}(\im{Z}_f,x,y)$ circularly translates $\im{Z}_f$ \wrt $(x,y)$ pixels, as defined in Eq. \eqref{eq.trans}.
\begin{equation}\label{eq.trans}
\textrm{trans}(\im{Z}_f,x,y)[:,i,j]=\im{Z}_f[:,\textrm{mod}(i+x,16),\textrm{mod}(j+y,16)]
\end{equation}
In \emph{Correspondence Match Pooling}, we extract out meaningful response maps using three types of pooling---average pooling as defined in Eq.~\eqref{eq.avgPool}
\begin{equation}\label{eq.avgPool}
\textrm{avgPool}\big( \textrm{corr}(\im{P}_f\!,\!\im{Q}_f)\big)[\!i\!,\!j]
\!\!\!=\!\!\!\textstyle\sum_{x=0}^{15}\sum_{y=0}^{15}\textrm{corr}(\im{P}_f\!,\!\im{Q}_f)[\!x,\!y,\!i,\!j]/256
\end{equation}
max pooling as defined in Eq.~\eqref{eq.maxPool}
\begin{equation}\label{eq.maxPool}
\textrm{maxPool}\big( \textrm{corr}(\im{P}_f,\im{Q}_f)\big)[i,j]=\max_{x,y}\{\textrm{corr}(\im{P}_f,\im{Q}_f)[x,y,i,j]\}
\end{equation}
and argsort pooling as defined in Eq.~\eqref{eq.argsortPool}
\begin{equation}\label{eq.argsortPool}
\textrm{argsortPool}\big( \textrm{corr}(\im{P}_f,\im{Q}_f)\big)[k]=\textrm{corr}(\im{P}_f,\im{Q}_f)[k_x,k_y,i,j]
\end{equation}
where $(k_x,k_y)$ in Eq. \eqref{eq.argsortPool} is determined by the $k$th maximum response over all translations.  Finally, we concatenate one average, one max, and the top six argsort response maps along the feature dimension and obtain the dense matching response $\im{Q}_r$ of shape $8\times 16 \times 16$ for $\im{Q}$. By interchanging the roles of $\im{P}_f$ and $\im{Q}_f$ in $\textrm{corr}(\cdot,\cdot)$, one can therefore obtain $\im{P}_r$.

Fig.~\ref{fig.visPooling} visualizes intermediate results of the proposed \emph{Deep Dense Matching} layer for the three testing samples from Fig.~\ref{fig.samples}. As one can see, the proposed deep dense matching module 1) successfully localizes potential splicing regions, and 2) produces substantially higher responses to the two positive samples above than the negative sample (see the red dash line on the right half stats figure in Fig.~\ref{fig.visPooling}), implying that the previous \emph{CNN Feature Extractor} and \emph{Deep Feature Correlation} provides meaningful representation with high discernibility. 

\begin{figure}[!h]
\scriptsize
\centering
\begin{tabular}{cc}
\includegraphics[width=.475\linewidth]{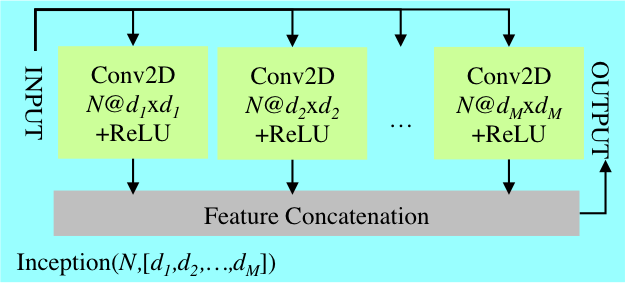}&
\includegraphics[width=.475\linewidth]{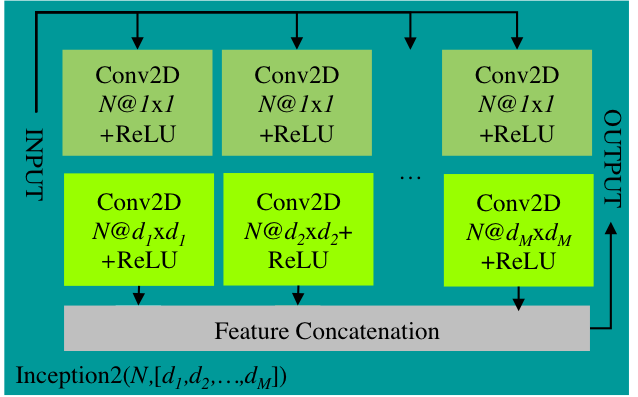}\\
\end{tabular}
\caption{The internal architecture of the two types of used inception modules.}\label{fig.inception}
\end{figure}

In order to produce a splicing mask from the dense response map, we use a \emph{Mask Deconvolution} module, as shown in Fig ~\ref{fig.pipeline}(b), where we gradually deconvolve a response map by a factor of 2 until its size reaches the size of input, i.e., $256\times 256$. During each deconvolution stage, we apply an inception module \cite{inception} with a larger filter size and a smaller number of filters, where the two types of inception modules can be seen in Fig.~\ref{fig.inception}. This enables us to obtain splicing masks for both image $\im{P}$ and $\im{Q}$, i.e., outputs $\im{P}_m$ and $\im{Q}_m$ (see examples in the ``Predicted Mask'' column in Fig.~\ref{fig.visPooling}.)

\subsection{Splicing Detection}
\begin{figure}[!h]
\scriptsize
\centering
\includegraphics[width=.65\linewidth]{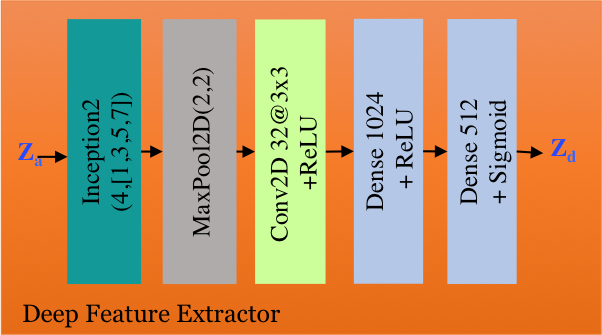}
\caption{Internal architecture of the deep feature extractor.}\label{fig.deepFeatex}
\end{figure}

Intuitively, given a query image $\im{Q}$ and a potential donor $\im{P}$ along with predicted splicing masks $\im{Q}_m$ and $\im{P}_m$, one can easily validate whether two masks match each other by their image contents. We therefore follow this intuition to design a \emph{Visual Consistency Validation} module to fulfill the splicing detection task as shown in Fig.~\ref{fig.pipeline}(b). Specifically, we first use the \emph{Visual Attention} module to zero-out all non-spliced regions in the CNN feature, as shown in Eq. \eqref{eq.att},
\begin{equation}\label{eq.att}
\textrm{visAtt}(\im{Z}_f,\im{Z}'_m)[c,i,j]=\left\{ 
\begin{array}{rl}
\im{Z}_f[c,i,j],&\textrm{if}\,\im{Z}'_m[i,j] > 0.5 \\
0,&\textrm{otherwise}\\
\end{array}
\right.
\end{equation}
where $\im{Z}'_m$ is the result of $\im{Z}_m$ after classic \emph{MaxPooling2D} for a size (16,16). This process is analogous to forcing the network to pay attention only to splicing regions. Furthermore, we follow a Siamese-like network to compare these two attention features---namely, extract a new round of features from the two attention features using the \emph{Deep Feature Extractor} (see Fig.~\ref{fig.deepFeatex} for detailed architecture), and then compute the difference between the two resulting features. We then concatenate this feature with the feature obtained from the average mask responses. Finally, we use three stacked dense layers to infer the probability that $p=\textrm{Proba}(\im{P} \,\textrm{is a donor of}\,\im{Q} )$, and this fulfills the detection task as shown in Fig.~\ref{fig.pipeline}(b).

\subsection{Training Data and Strategy}
To the best of our knowledge, no dataset exists that is large enough to be directly used for training the proposed DMVN model. To overcome this limitation, we use the SUN2012 object detection dataset~\cite{sun2012} and the MS COCO dataset~\cite{coco} to create training samples according to the unsupervised generation process described in \cite{zhu2015learning}. Briefly, we begin with a random image $\im{X}$ with polygon-based object annotations, randomly select an object in $\im{X}$, then randomly transform this object and paste it to another randomly selected image $\im{Y}$ to obtain a resulting composite image $\im{Z}$. We could harvest at most three (two positive and one negative) training $\{inputs,outputs\}$ samples for each unsupervised data generation. For instance, Fig.~\ref{fig.samples} gives a set of three training samples of this type.

In terms of the parameters controlling the data generation process, we equally likely pick an image and an object, random affine transformation involving a scale change in $\mathbb{U}(.5,4)$, rotation in $\mathbb{U}(-10,+10)$, shift in $\mathbb{U}(-127,+127)$, translation $\mathbb{U}(-127,+127)$ and random luminance change in $\mathbb{U}(-32,+32)$. This enables us to create as many samples as needed to train the network end-to-end. Effectively, we create 1.5 M(illion), 0.3M, and 0.3M synthesized samples for training, validation, and testing, respectively.

The proposed DMVN was implemented using the \emph{Keras} deep learning framework with the \emph{Theano} backend, including all custom correlation and pooling layers. Our model was trained with the \emph{Adadelta} optimizer \wrt the \emph{log loss} for both localization and detection branches. Since we design the splicing detection branch as a \emph{Visual Consistency Validator} of image contents on predicted splicing masks, this branch output may not produce meaningful gradients unless the localization branch produced meaningful splicing masks. Thus, we first focus on the localization branch of DMVN model only. Once this localization branch converges, we freeze its weights, add on the detection branch, and train the detection branch until it converges. We finally unfreeze all DMVN weights and train the entire model end-to-end using the stochastic gradient optimizer with a learning rate $1e-5$ and momentum of 0.9. In summary, we achieve 98.52\%, 98.67\%, and 97.88\% prediction accuracy on the localization branch, and 97.75\%, 97.53\% and 97.69\% prediction accuracy on the detection branch on our synthesized training, validation, and testing datasets, respectively. Our pretrained model can be downloaded from \url{https://gitlab.com/rex-yue-wu/Deep-Matching-Validation-Network.git}.

\section{Experimental Results}\label{sec.experiments}
\subsection{Baseline Methods and Test Settings}
Since the CISD formulation is completely new, we compare against baseline algorithms from the state-of-the-art copy-move detection algorithms. We rely on visual clues, because when we concatenate the two inputs from a CISD sample into a single combined image, the resulting image contains copy-move forgery if the CISD sample is positive. Specifically, we choose the classic block matching-based approach \cite{luo2006robust}, the classic Zernike moments-based block matching \cite{ryu2010detection} with nearest-neighbor search, the SURF feature-based keypoint matching \cite{christlein2012evaluation} and the dense field matching \cite{cozzolino2015efficient}. All used baselines are implemented by either a third-party or by the authors of the corresponding papers.\footnote{Available at \url{https://github.com/rahmatnazali/image-copy-move-detection.git}, \url{https://www5.cs.fau.de/research/software/copy-move-forgery-detection/}, \url{http://www.grip.unina.it/research/83-image-forensics/90-copy-move-forgery.html} as the date of April 10, 2017.}

With regard to preprocessing, we resize an image to $256\times 256$ to fit the input size of the proposed DMVN, and thus a $256\times 512$ image for those baseline algorithms. With regard to postprocessing, we do not apply any to DMVN, i.e., using the outputs from DMVN localization and detection branches directly, while keeping default postprocessing settings of baselines unchanged. Since some baseline methods only output a splicing mask but not a binary decision on detection, we follow the tradition in the classic ISD and copy-move community to determine that a sample is positive if no pixel is labeled as spliced in a mask. Finally, all baseline methods are run on \texttt{Intel Xeon} CPU E5-2695@2.40GHz, and the proposed DMVN is run on \texttt{Nividia TitanX} GPU.  

\subsection{Dataset}
We conduct evaluation experiments on two large datasets: 1) the paired CASIA dataset, and 2) the NIST-provided Nimble 2017 image splicing detection dataset. The paired CASIA dataset is a modified version of the original CASIA TIDEv2.0 dataset~\cite{wang2010image}\footnote{\url{http://forensics.idealtest.org/casiav2/}}
which contains 7200 authentic color images and 5123 color images tampered with by realistic manual manipulations (e.g., resize, deform, and  blurring) through \emph{Adobe Photoshop CS3}. It was originally collected for both the image copy-move problem and the classic image splicing detection problem. Since the CISD problem requires a pair of inputs, we select pairs of images from the original CASIA dataset to create the new {paired CASIA dataset}. Among the 5123 CASIA tampered images, we that found 3302 are of the {copy-move} problem and 1821 are of the classic ISD task. We therefore generate 3642 positive samples by pairing these 1821 spliced images with their true donor images, and collect 5000 negative samples by randomly pairing 7491 color images from the same CASIA-defined content category. Our paired CASIA dataset can be found at \url{https://gitlab.com/rex-yue-wu/Deep-Matching-Validation-Network.git}.

With regard to the Nimble 2017 dataset, it is provided by NIST with 98 positive samples and 529,836 negative samples. This challenging dataset is particularly designed for the CISD task with considerations to 1) a very large scale (more than a half million samples), 2) more realistic and artistic manipulations like image impainting, seam-carving etc., 3) difficult negative samples with visually similar foreground and background, and 4) mimicking the real application scenario where the ratio of negative samples to positive samples is extremely huge.

It is worth emphasizing that 1) we directly test the DMVN models trained by our synthetic data without any finetuning, and 2) ground truth splicing masks are not available for both dataset. 
\subsection{Evaluation Metrics}
Since both datasets do not provide ground truth splicing masks, we focus on assessing the splicing detection performance. We follow the tradition of the classic ISD and copy-move community, namely, using precision, recall, and f-score: TP stands for \textit{true positive}, i.e., correctly detected as spliced; FN stands for\textit{ false negative}, i.e., incorrectly detected as not-spliced; FP stands for \textit{false positive}, i.e., incorrectly detected as spliced; and TN stands for \textit{true negative}, i.e., correctly detected as not-spliced. 
\begin{equation}\label{eq.precision}
\textrm{precision} = TP/(TP+FP)
\end{equation}
\begin{equation}\label{eq.recall}
\textrm{recall} = TP/(TP+FN)
\end{equation}
\begin{equation}\label{eq.fscore}
\textrm{f-score} = 2TP/(2TP+FN+FP)
\end{equation}
Furthermore, we also use \textit{area under the ROC curve (AUC) }to evaluate overall system performance at different operation points, where the receiver operating characteristic (ROC) curve is determined as the function of \textit{true positive rate} \textit{(TPR)} in terms of \textit{false positive rate (FPR)}. TPR and FPR are defined as shown in Eqs.~\eqref{eq.TPR} and ~\eqref{eq.FPR}.  The area under an ROC curve then quantifies the over-ability of the system to discriminate between two classes. It is worth noting that a system which is no better at identifying true positives than random guessing has an area of 0.5, and a perfect system without false positives and false negatives has an area of 1.0; and that AUC is the only official metric used by the Nimble 2017 challenge .
\begin{equation}\label{eq.TPR}
TPR=TP/(TP+FN)
\end{equation}
\begin{equation}\label{eq.FPR}
FPR=FP/(TN+FP)
\end{equation}

\subsection{Results}
The most challenging task in the CISD is to 1) find spliced regions under various transformations like translation, rotation, scale, crop, etc., while dealing with complicated cases like multiple instances (a donor image contains multiple instances that are similar to a true spliced region), and multiple spliced regions (a donor image contributes more than one region); and 2) reduce false alarms on those visually similar but non-spliced regions. 

\def\tww{.9cm}
\begin{table}[!h]
\scriptsize
\centering
\caption{CISD performance comparison on CASIA }\label{tab.casia}
\begin{tabular}{@{}R{1cm}|C{\tww}@{}C{\tww}@{}C{\tww}|L{2cm}@{}}
\hline
\textbf{Method}&\textbf{Precision}&\textbf{Recall}&\textbf{F-score}&\textbf{Time (sec/sample)}\\\hline
\cite{christlein2012evaluation}&51.64\%&82.92\%&63.64\%&1.85\\
\cite{luo2006robust}&99.69\%&53.53\%&69.66\%&6.27$\times 10^{+2}$\\
\cite{ryu2010detection}&96.14\%&58.95\%&73.09\%&8.61\\
\cite{cozzolino2015efficient}&98.97\%&63.34\%&77.25\%&3.23\\\hline
DMVN loc.&91.52\%&79.18\%&84.91\%&7.16$\times 10^{-2}$\\
DMVN det&94.15\%&79.08\%&85.96\%&8.29$\times 10^{-2}$\\
\hline
\end{tabular}
\end{table}

\begin{figure}[!h]
\centering
\scriptsize
\begin{tabular}{c@{}c}
\includegraphics[trim =0cm 0cm 0cm .2cm,clip, width=.475\linewidth]{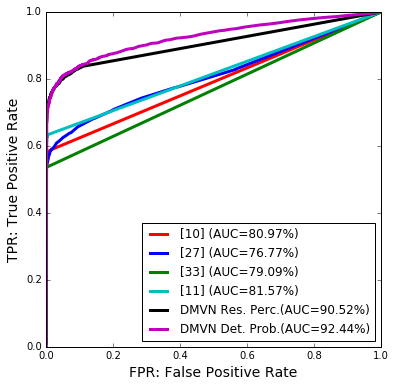}&
\includegraphics[trim =0cm 0cm 0cm .2cm,clip, width=.475\linewidth]{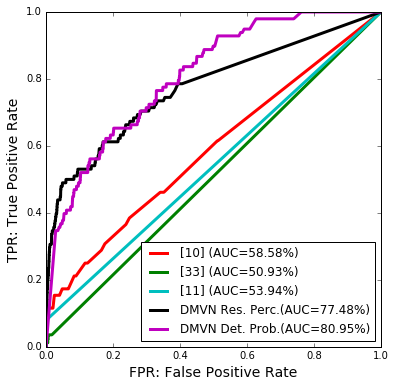}\\
(a) ROC on the paired CASIA dataset& (b) ROC on the Nimble 2017 dataset\\
\end{tabular}
\caption{CISD performance comparison using ROC.}\label{fig.auc_casia}
\end{figure}

Table \ref{tab.casia} shows the splicing detection performance of baseline approaches and the proposed DMVN methods on the paired CASIA dataset, where \emph{DMVN loc.} means that we determine whether a sample (a pair of images) is positive or not by checking whether any pixel in a predicted splicing mask (from the DMVN localization branch) is positive, and \emph{DMVN det.} means that we directly use the relation probability (from the DMVN detection branch) to determine a sample's label. As one can see, the proposed DMVN methods outperform peer algorithms by a large margin in terms of f-score ($\sim$7\% higher) on the paired CASIA dataset. Note also that the proposed DMVN is significantly faster than baseline approaches by 20+ times, and that the DMVN detection branch improves our precision score from 91.52\% to 94.15\% while only reducing recall score for 0.1\%, thus indicating the effectiveness of the proposed validation idea which relies on visual attention and Siamese architecture. Fig.~\ref{fig.auc_casia} compares ROC and AUC scores for different methods, where \emph{DMVN Res. Perc.} and \emph{DMVN Det. Prob.} means that the threshold used to obtain TPR and FPR is based on the positive pixel percentage in a resulting mask, and the detection branch's output probability, respectively. Again, the proposed DMVN methods are noticeably better than others on AUC scores (10\%+ on CASIA, and 20\%+ on Nimble 2017).

\begin{figure*}[tbph]
\centering
\scriptsize
\begin{tabular}{C{\tw}@{}C{\tw}@{}C{\tw}@{}C{\tw}@{}C{\tw}@{}C{\tw}@{}C{.01cm}@{}C{\tw}@{}C{\tw}@{}C{\tw}@{}C{\tw}@{}}
\multicolumn{1}{c}{\sp$\im{P}$\sp}&\multicolumn{1}{c}{\sp$\im{Q}$\sp}&\multicolumn{1}{c}{\spb$\im{P}_m$\sp}&\multicolumn{1}{c}{\spb$\im{Q}_m$\spb}&\multicolumn{1}{c}{\spc$\im{P}_m * \im{P}$\spc}&\multicolumn{1}{c}{\spc$\im{Q}_m*\im{Q}$\spc}&&\multicolumn{1}{c}{\sp\spc$\im{P}$\sp}&\multicolumn{1}{c}{\sp$\im{Q}$\sp}&\multicolumn{1}{c}{\sp$\im{P}_m$\sp}&\multicolumn{1}{c}{\sp$\im{Q}_m$\sp}\\
\multicolumn{6}{c}{\includegraphics[width=.581\linewidth]{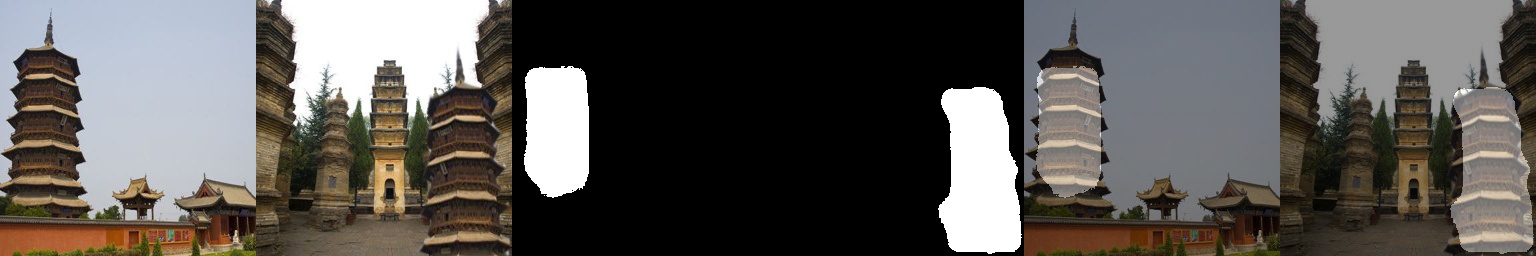}}&&
\multicolumn{4}{c}{\includegraphics[width=.385\linewidth]{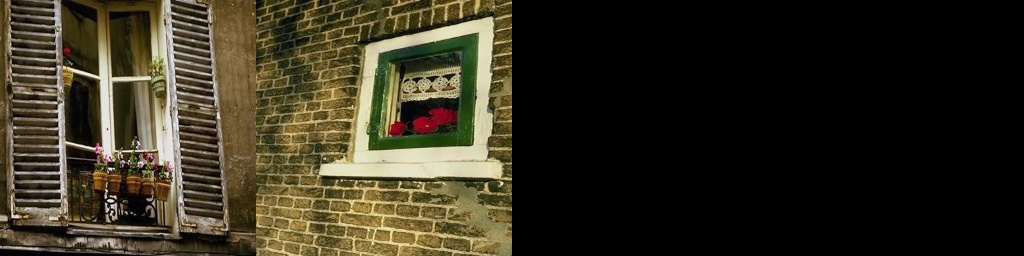}}\\
\multicolumn{6}{c}{(PID, QID)=(Au\_arc\_00072, Tp\_D\_CRN\_M\_N\_arc00073\_arc00072\_10267) \simiFg, \translation, \scale, \rotation, \multiInst} && \multicolumn{4}{c}{(PID, QID)=(Au\_art\_30406, Au\_art\_30197) \simiFg, \simiBg}\\
\multicolumn{6}{c}{\includegraphics[width=.581\linewidth]{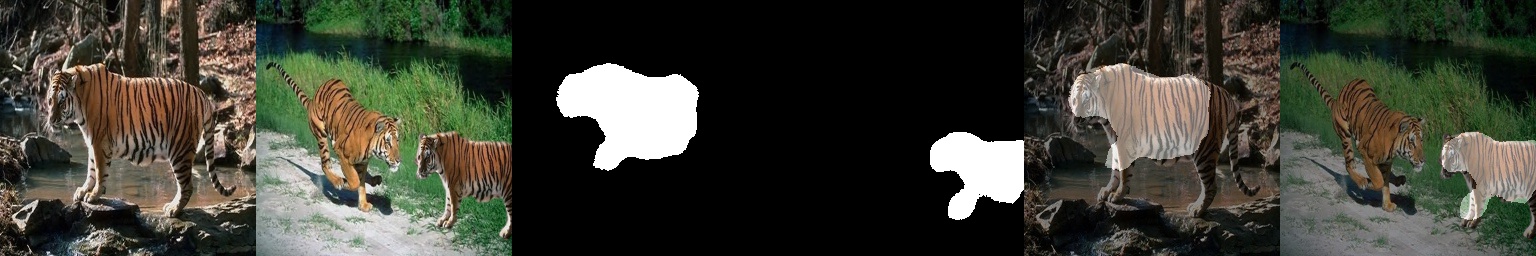}}&&
\multicolumn{4}{c}{\includegraphics[width=.385\linewidth]{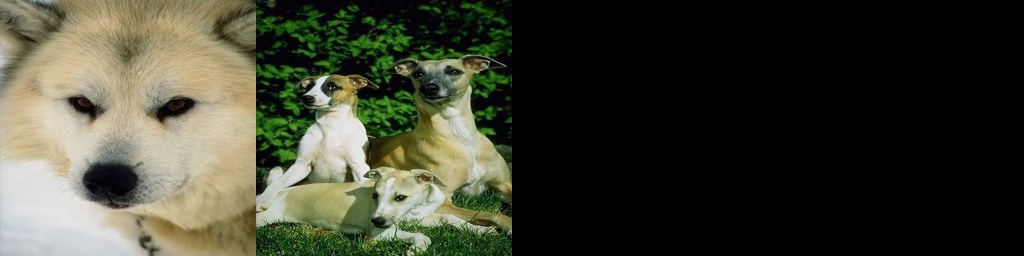}}\\
\multicolumn{6}{c}{(PID, QID)=(Au\_ani\_00037,Tp\_D\_NRN\_S\_B\_ani00036\_ani00037\_00156) \simiFg, \translation, \scale, \matchPart, \multiInst} && \multicolumn{4}{c}{(Au\_ani\_30707,Au\_ani\_30697) \simiFg, \multiInst}\\
\multicolumn{6}{c}{\includegraphics[width=.581\linewidth]{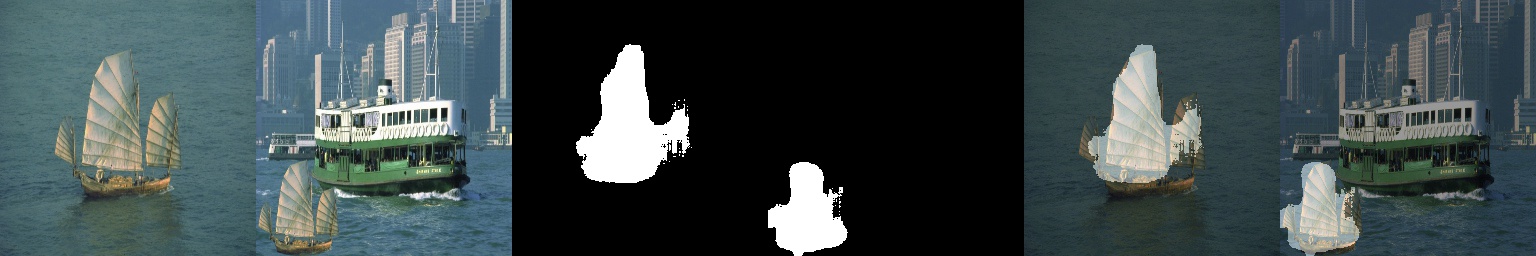}}&&
\multicolumn{4}{c}{\includegraphics[width=.385\linewidth]{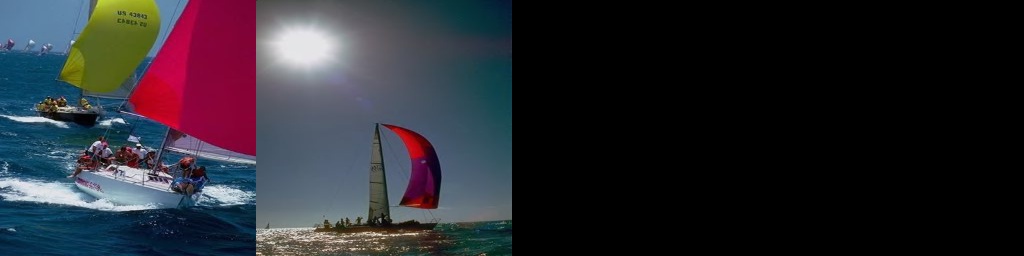}}\\
\multicolumn{6}{c}{(PID, QID)=(Au\_art\_20005,Tp\_D\_NRN\_S\_N\_art20006\_art20005\_01812) \simiFg, \simiBg, \translation, \scale, } && \multicolumn{4}{c}{(PID, QID)=(Au\_sec\_30528,Au\_sec\_30506) \simiFg, \simiBg, \multiInst}\\
\multicolumn{6}{c}{\includegraphics[width=.581\linewidth]{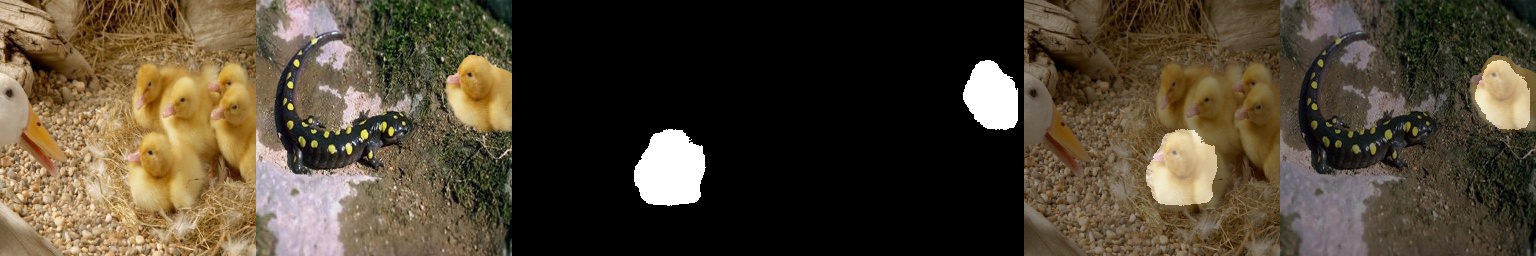}}&&
\multicolumn{4}{c}{\includegraphics[width=.385\linewidth]{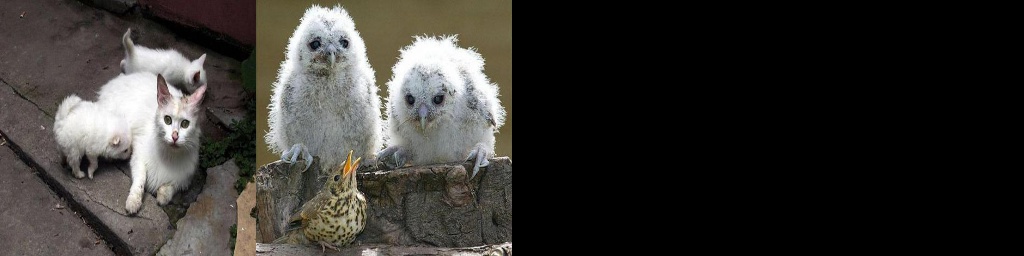}}\\
\multicolumn{6}{c}{(PID, QID)=(Au\_ani\_00089,Tp\_D\_NNN\_S\_B\_ani00055\_ani00089\_11137) \simiFg, \simiBg, \translation,  \matchPart, \multiInst} && \multicolumn{4}{c}{(PID, QID)=(Au\_ani\_20012,Au\_ani\_20009) \simiFg, \simiBg, \multiInst}\\
\multicolumn{6}{c}{\includegraphics[width=.581\linewidth]{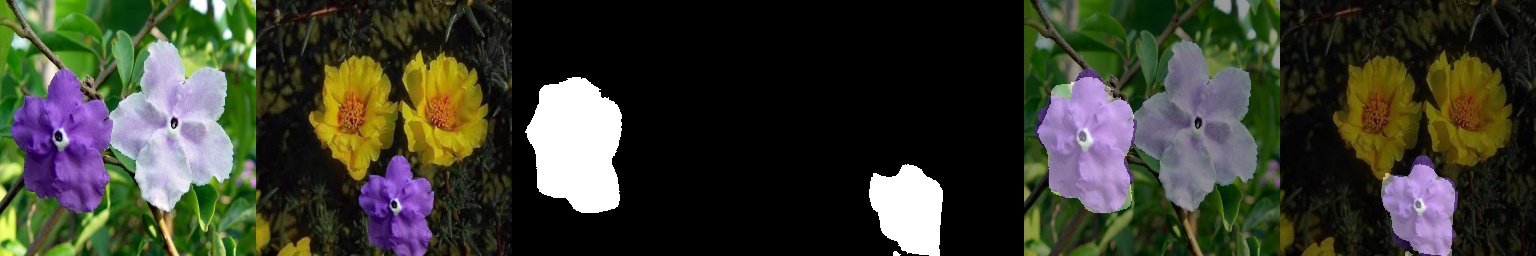}}&&
\multicolumn{4}{c}{\includegraphics[width=.385\linewidth]{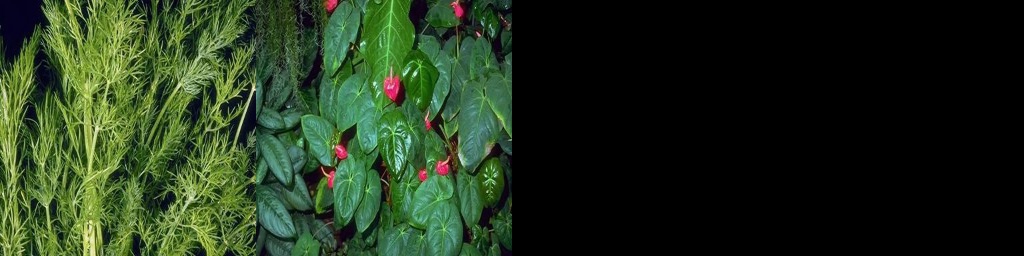}}\\
\multicolumn{6}{c}{(PID, QID)=(Au\_pla\_10126,Tp\_D\_NRN\_S\_N\_pla00020\_pla10126\_12131) \simiFg, \simiBg, \translation, \scale, \multiInst} && \multicolumn{4}{c}{(PID, QID)=(Au\_pla\_30703,Au\_pla\_00045) \simiFg, \simiBg}\\
\multicolumn{6}{c}{\includegraphics[width=.581\linewidth]{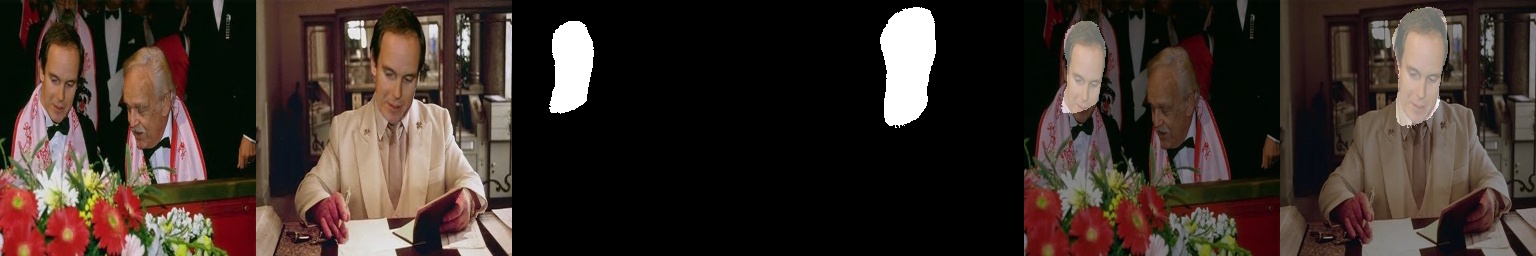}}&&
\multicolumn{4}{c}{\includegraphics[width=.385\linewidth]{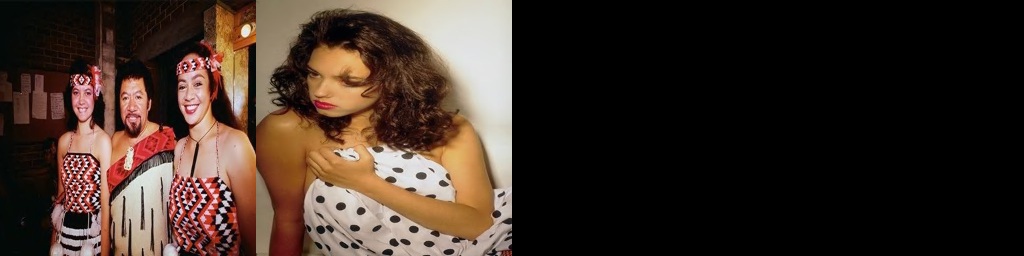}}\\
\multicolumn{6}{c}{(PID, QID)=(Au\_ind\_00025,Tp\_D\_NRN\_S\_N\_ind00021\_ind00025\_10398) \simiFg, \simiBg, \translation, \scale, \multiInst }&&
\multicolumn{4}{c}{(PID, QID)=(Au\_cha\_30609,Au\_cha\_30425) \simiFg, \simiBg, \multiInst}\\
\multicolumn{6}{c}{\includegraphics[width=.581\linewidth]{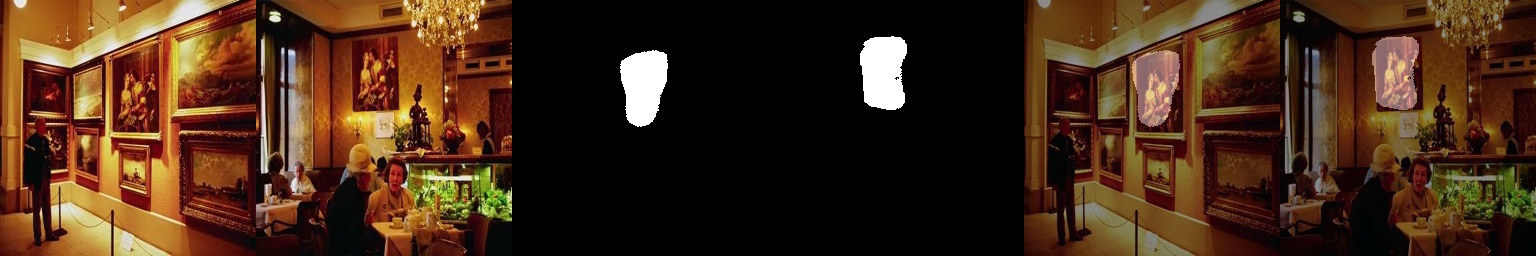}}&&
\multicolumn{4}{c}{\includegraphics[width=.385\linewidth]{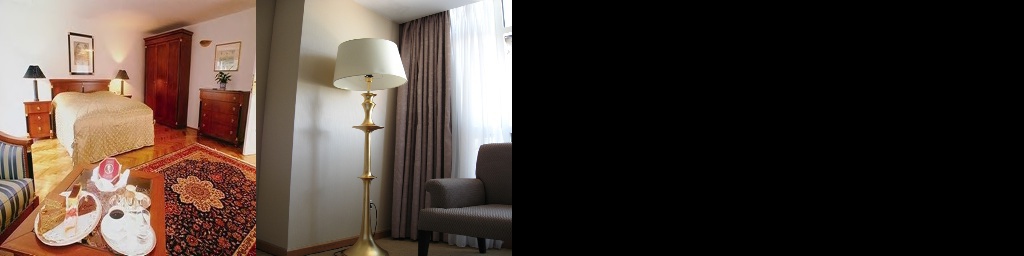}}\\
\multicolumn{6}{c}{(PID, QID)=(Au\_ind\_00077,Tp\_D\_CND\_S\_N\_ind00078\_ind00077\_00476) \simiFg, \simiBg, \translation, \scale, \rotation, \perspective }&&
\multicolumn{4}{c}{(PID, QID)=(Au\_ind\_00084,Au\_ind\_10002) \simiFg, \simiBg}\\
\multicolumn{6}{c}{\includegraphics[width=.581\linewidth]{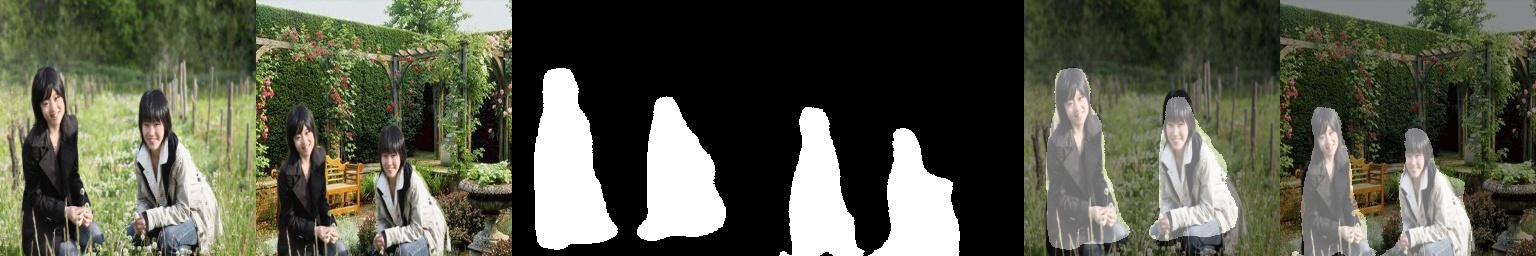}}&&
\multicolumn{4}{c}{\includegraphics[width=.385\linewidth]{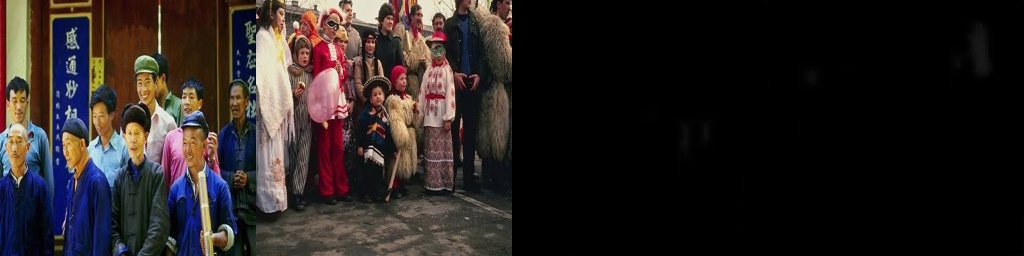}}\\
\multicolumn{6}{c}{(PID, QID)=(Au\_cha\_00095,Tp\_D\_NRN\_M\_B\_sec00020\_cha00095\_00041) \simiFg, \simiBg, \translation, \scale, \multiSpli, \multiInst}&&
\multicolumn{4}{c}{(PID, QID)=(Au\_sec\_30331,Au\_sec\_30019) \simiFg, \simiBg, \multiInst}\\
\multicolumn{6}{c}{\includegraphics[width=.581\linewidth]{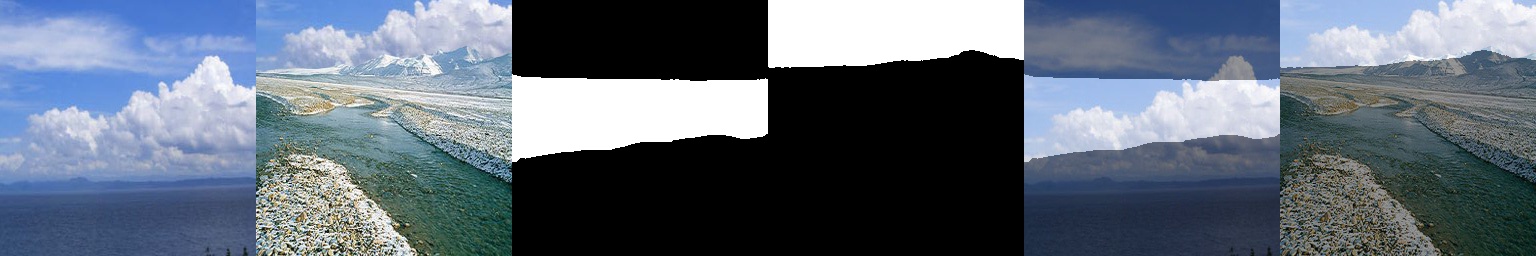}}&&
\multicolumn{4}{c}{\includegraphics[width=.385\linewidth]{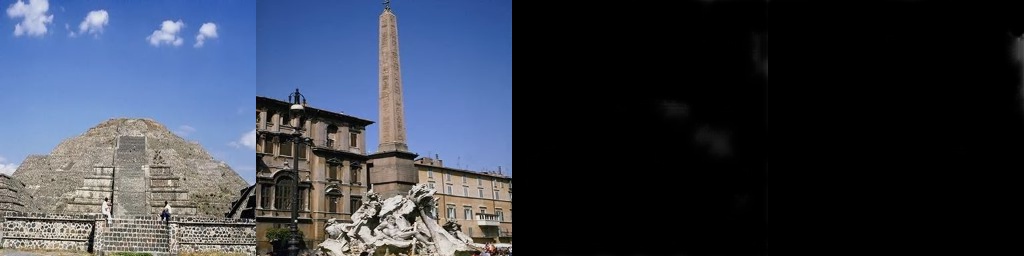}}\\
\multicolumn{6}{c}{(PID, QID)=(Au\_nat\_00097,Tp\_D\_NNN\_M\_N\_nat00075\_nat00097\_11100) \simiFg, \simiBg, \translation, \matchPart}&&
\multicolumn{4}{c}{(PID, QID)=(Au\_arc\_30072,Au\_arc\_30435) \simiFg, \simiBg}\\
\multicolumn{6}{c}{\includegraphics[width=.581\linewidth]{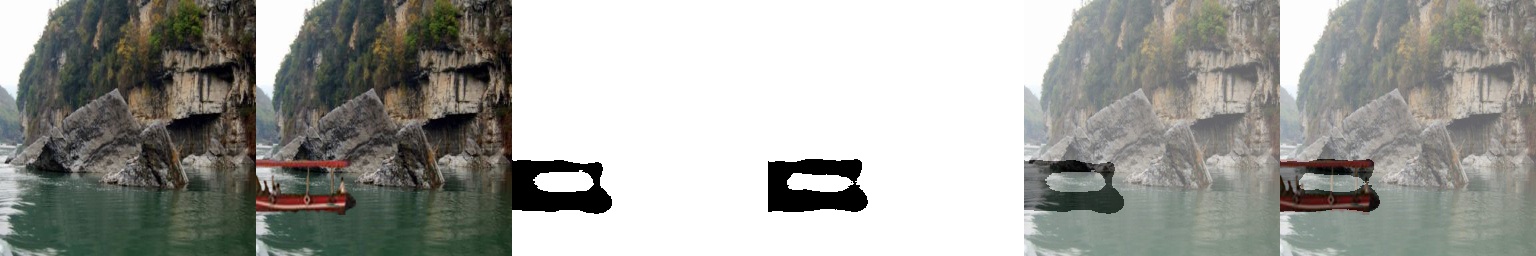}}&&
\multicolumn{4}{c}{\includegraphics[width=.385\linewidth]{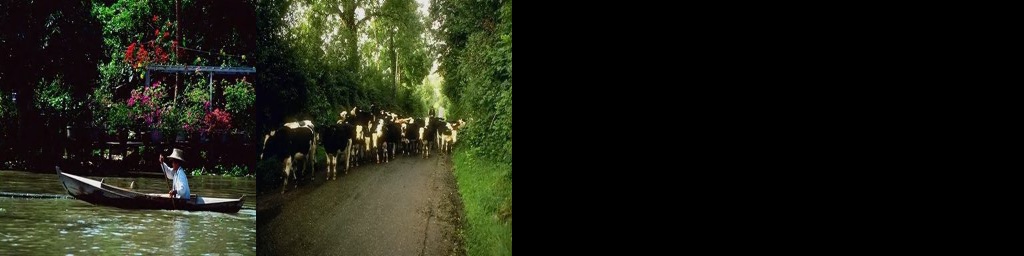}}\\
\multicolumn{6}{c}{(PID, QID)=(Au\_sec\_00085,Tp\_D\_CRN\_M\_B\_sec00085\_arc00065\_11450) \simiFg, \simiBg}&&
\multicolumn{4}{c}{(PID, QID)=(Au\_sec\_30331,Au\_sec\_30019) \simiFg, \simiBg}\\
\multicolumn{6}{c}{(a) True positive samples}&&\multicolumn{4}{c}{(b) True negative samples}\\
\end{tabular}
\caption{DMVN localized splicing masks on the paired CASIA TIDEv2.0 dataset. (PID, QID) indicate the original CASIA filename of $\im{P}$ and $\im{Q}$. Color blocks indicate different factors which splicing localization need to be robust against, namely \translation: translation, \scale: scale, \rotation: rotation, \perspective: perspective, \matchPart: crop, \multiInst: multiple instances, \multiSpli: multiple splicing objects,  \simiFg:similar foreground, \simiBg: similar background.}\label{fig.visCASIA}
\end{figure*}

\begin{figure}[!h]
\centering
\scriptsize
\begin{tabular}{@{}c c@{}}
\includegraphics[width=.4\linewidth]{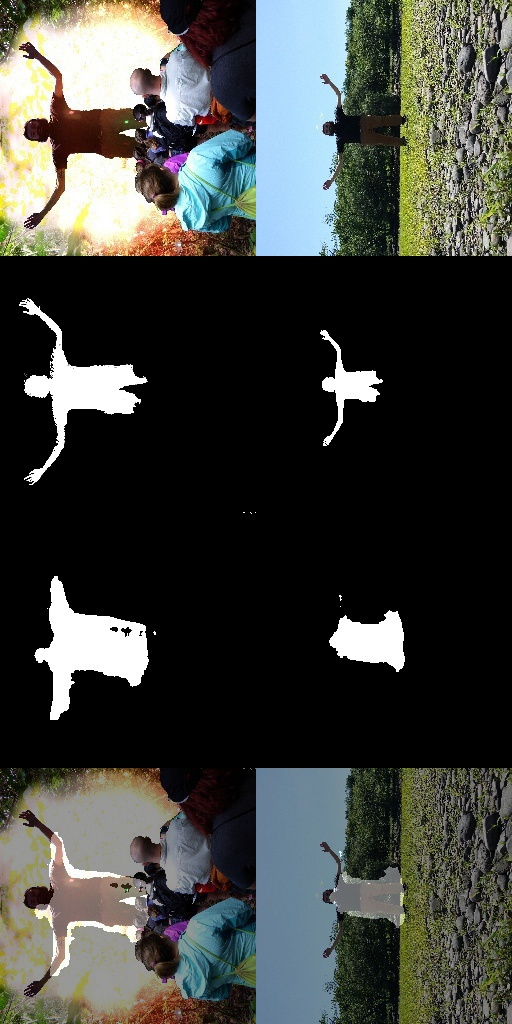}&
\includegraphics[width=.4\linewidth]{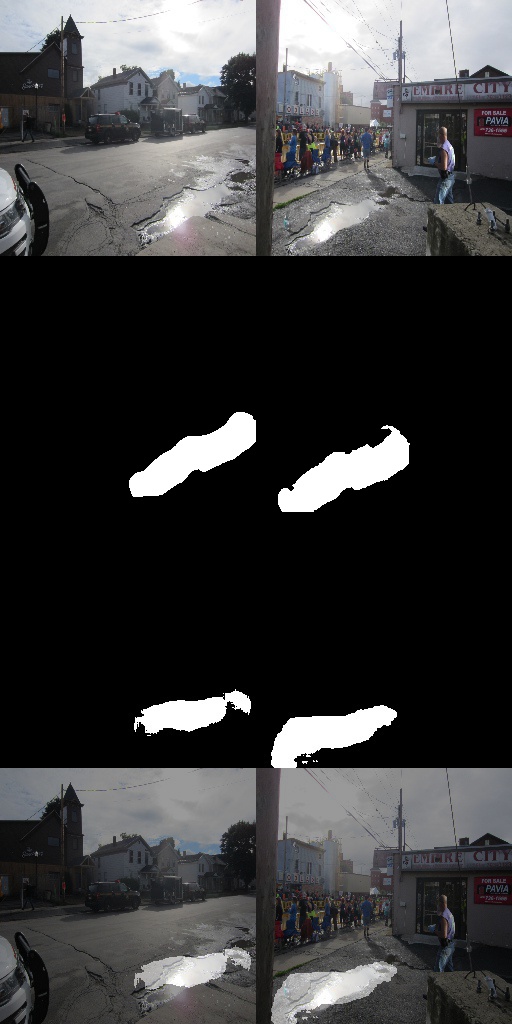}\\
(a) (PID,QID)=&(b) (PID,QID)=\\
(19e32f2b29a2b345dfd28a8ad857b3fc,&(c0cfa4a6a95b3762f9d966e09a0340b9,\\
797a0b5dd9c47a4b391f4cee60ba9354)&030d521fca99981c442e60e4b2155699)\\
\includegraphics[width=.4\linewidth]{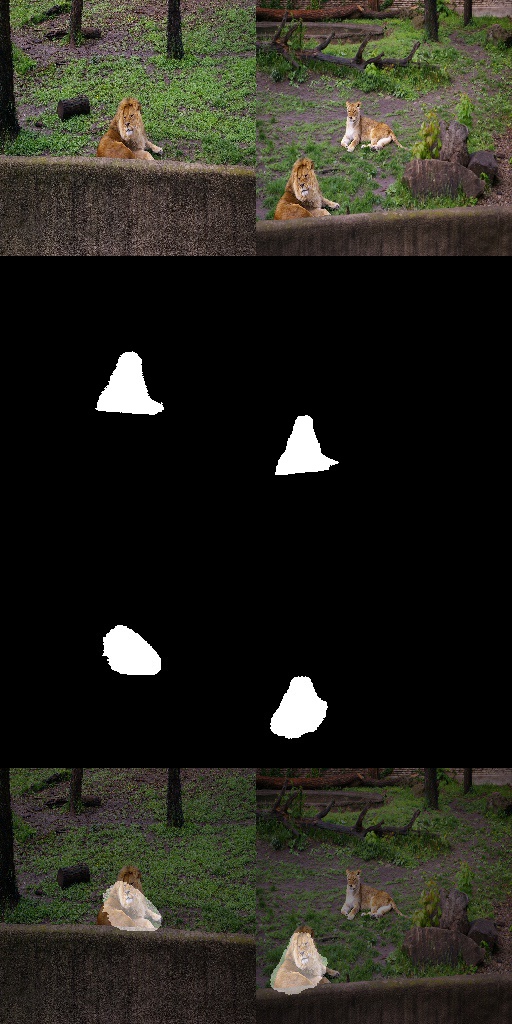}&
\includegraphics[width=.4\linewidth]{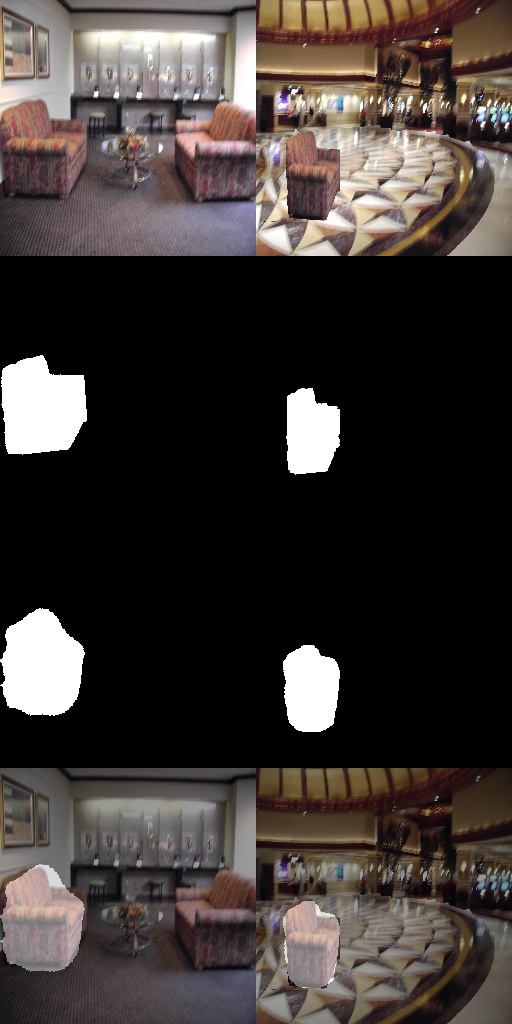}\\
(c) (PID,QID)=&(d) (PID,QID)=\\
(9883d12156ed82344d65fe19575e1d84,&(4bf6a60d74ecd725e3e4b0eedc7d7cc0,\\
8b570c4bd216f5144774b9c0bc61bbd4)&96492c250780ebababfcebc585ef1db6)\\
\end{tabular}
\caption{Predicted splicing masks on NC2017 dataset. From top to bottom, input pair, manually annotated ground truth masks, predicted splicing masks, and overlaid masks.}\label{fig.visNC2017}
\end{figure}

With regard to splicing localization performance, Fig.~\ref{fig.visCASIA} shows how the proposed DMVN method conquered this challenge on the paired CASIA dataset, where $\im{X}_m$ indicates a splicing mask binarized with threshold 0.5, and $\im{X}_m*\im{X}$ indicates an overlaid image by using the splicing mask as the alpha channel with 40\% transparency. To see this, one shall go to each row in Fig.~\ref{fig.visCASIA}, where the left and right sides show true positive and negative samples, respectively. Note that two samples on each row are intentionally picked from a similar CASIA category and/or a similar object class. As one can see, the proposed DMVN method not only predicts meaningful splicing masks on those positive samples, but also correctly suppresses splicing masks on those negative samples. Fig.~\ref{fig.visNC2017} shows the manually annotated ground truth masks along with our predicted masks for the Nimble 2017 dataset.

\subsection{Discussions}
Fig.~\ref{fig.compareCasia} qualitatively compares the splicing localization performance for all supported baselines. As one can see: 1) classic exhaustive block matching method \cite{luo2006robust} is sensitive to transformation, but good at capturing nearly duplicate regions; 2)  block matching algorithm relying on Zernike moments \cite{ryu2010detection} handles a certain level of transformations, but fails to maintain the completeness of a splicing region (see those holes in ``\cite{ryu2010detection}'s Masks'' in row 5 of Fig.~\ref{fig.compareCasia}); 3) a keypoint-based detector \cite{christlein2012evaluation} may fail due to no effective keypoints or noisy keypoints, which can commonly be seen in a homogeneous region or regions with similar texture, and one has to further convert potential matching points to a mask (see the last row; finding correspondence does not mean find a mask); and 4) the proposed DMVN method does not suffer the drawbacks of the previous three methods and gives satisfactory localization results on homogeneous and non-homogeneous regions even under severe transformations. 

\notsure{With regard to drawbacks, the proposed DMVN has some difficulty detecting splicing objects smaller than $8\times 8$; this is due to the down-sampling in \emph{CNN Feature Extractor}. As one may notice, our AUC scores on the Nimble 2017 dataset are much lower than those on the paired CASIA dataset. Besides the fact of extremely unbalanced positive (98) and negative samples (529836), we notice that the Nimble 2017 dataset contains much more challenging samples. For example, the proposed DMVN approach produces false alarms when two images $\im{P}$ and $\im{Q}$ are from consecutive video frames, because it mistakenly predicts those similar but genuine objects in both $\im{P}$ and $\im{Q}$ as spliced objects.}

\def\fw{0.31}
\begin{figure}[!t]
\centering
\scriptsize
\begin{tabular}{@{}r@{}c@{}p{.05cm}@{}c@{}p{.05cm}@{}c@{}}
\raisebox{2\normalbaselineskip}[0pt][0pt]{\rot{Inputs}}&
\includegraphics[width=\fw\linewidth]{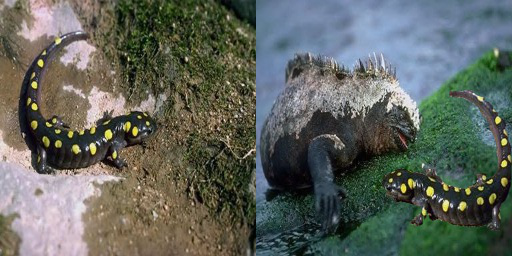}&&
\includegraphics[width=\fw\linewidth]{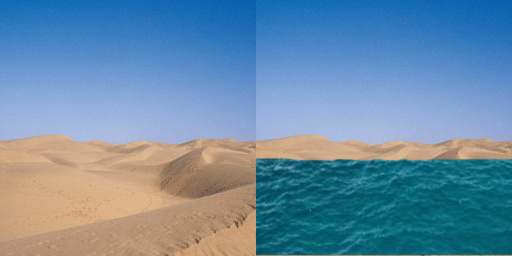}&&
\includegraphics[width=\fw\linewidth]{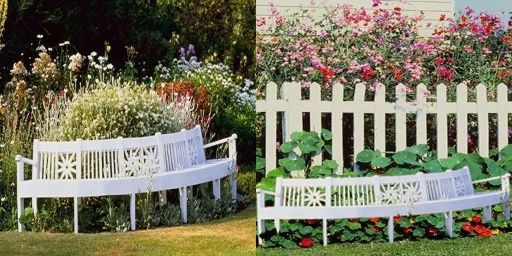}
\\\arrayrulecolor{red}\hline
\raisebox{2\normalbaselineskip}[0pt][0pt]{\rot{G.T. Masks}}&
\includegraphics[width=\fw\linewidth]{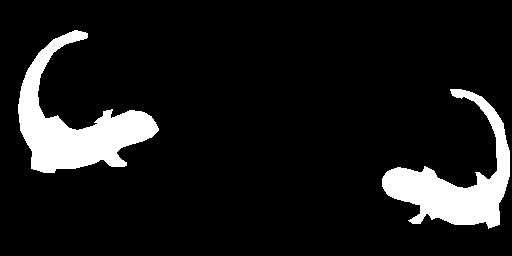}&&
\includegraphics[width=\fw\linewidth]{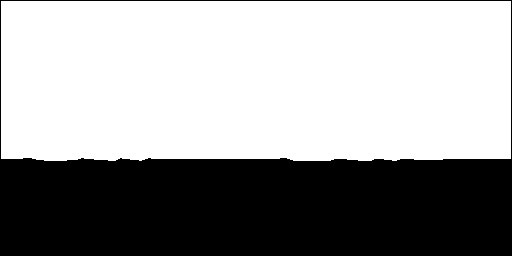}&&
\includegraphics[width=\fw\linewidth]{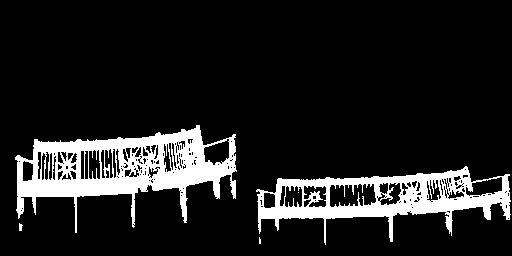}
\\\hline
\raisebox{2\normalbaselineskip}[0pt][0pt]{\rot{Ours Masks}}&
\includegraphics[width=\fw\linewidth]{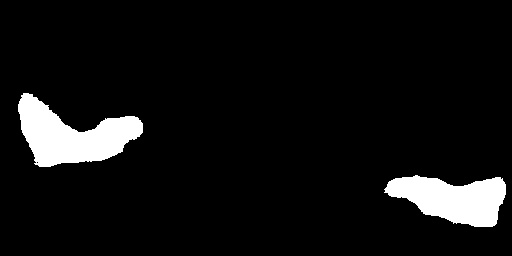}&&
\includegraphics[width=\fw\linewidth]{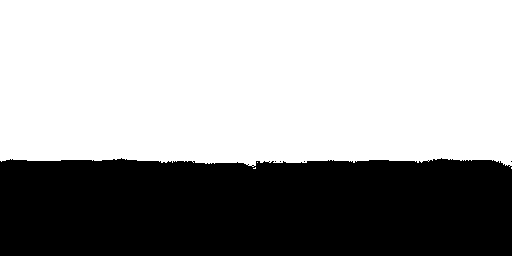}&&
\includegraphics[width=\fw\linewidth]{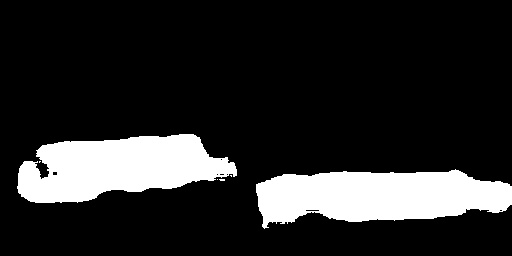}
\\\hline
\raisebox{2\normalbaselineskip}[0pt][0pt]{\rot{\cite{luo2006robust}'s Masks}}&
\includegraphics[width=\fw\linewidth]{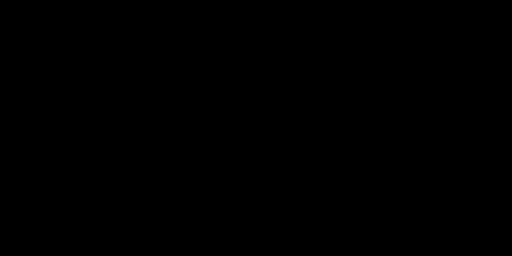}&&
\includegraphics[width=\fw\linewidth]{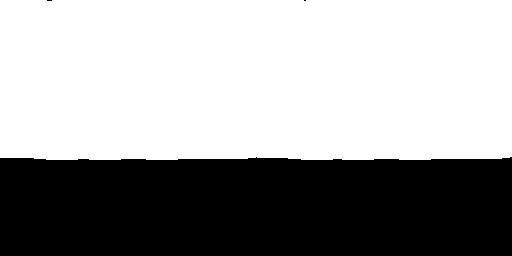}&&
\includegraphics[width=\fw\linewidth]{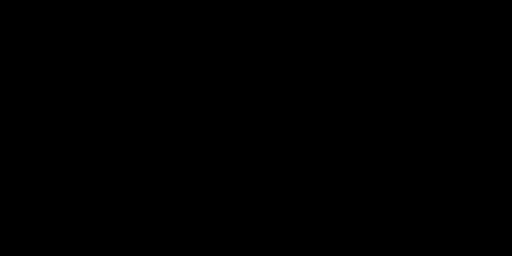}
\\\hline
\raisebox{2\normalbaselineskip}[0pt][0pt]{\rot{\cite{ryu2010detection}'s Masks}}&
\includegraphics[width=\fw\linewidth]{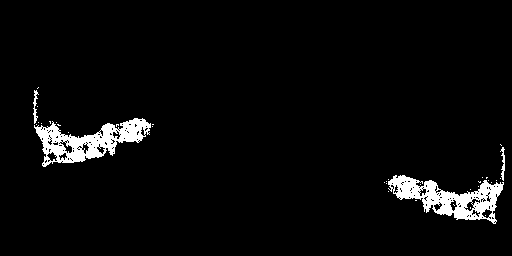}&&
\includegraphics[width=\fw\linewidth]{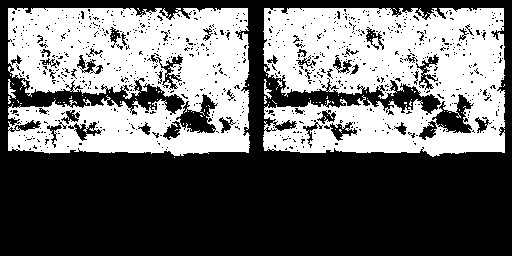}&&
\includegraphics[width=\fw\linewidth]{PR-Au_sec_00016_x_Tp_D_NRD_S_N_sec00011_sec00016_00031-0}
\\\hline
\raisebox{2\normalbaselineskip}[0pt][0pt]{\rot{\cite{christlein2012evaluation}'s Corr. Pts.}}&
\includegraphics[width=\fw\linewidth]{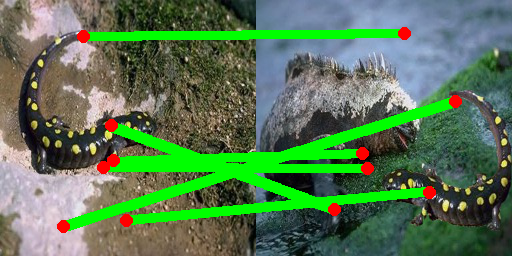}&&
\includegraphics[width=\fw\linewidth]{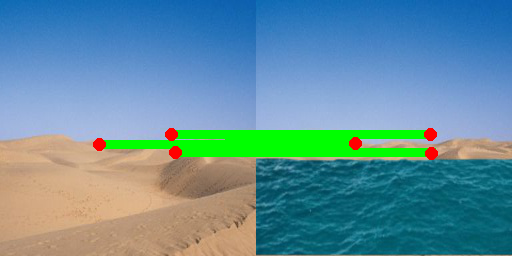}&&
\includegraphics[width=\fw\linewidth]{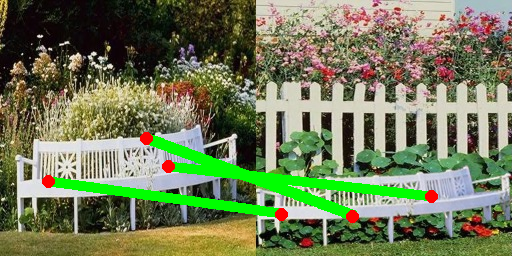}
\\
\raisebox{2\normalbaselineskip}[0pt][0pt]{\rot{\cite{christlein2012evaluation}'s Masks}}&
\includegraphics[width=\fw\linewidth]{PR-Au_ani_00055_x_Tp_D_CNN_M_N_ani00057_ani00055_11149}&&
\includegraphics[width=\fw\linewidth]{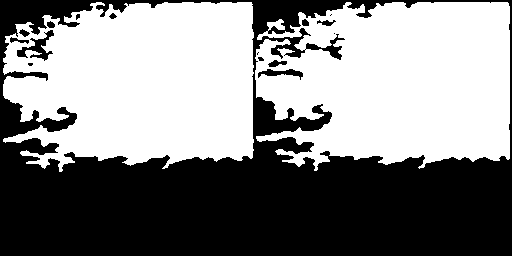}&&
\includegraphics[width=\fw\linewidth]{PR-Au_ani_00055_x_Tp_D_CNN_M_N_ani00057_ani00055_11149}
\\
&\tiny{(PID,QID)=(Au\_ani\_00055,Tp\_D\_}&
&\tiny{(PID,QID)=(Au\_nat\_00095,Tp\_D\_}&
&\tiny{(PID,QID)=(Au\_sec\_00016,Tp\_D\_}\\
&\tiny{CNN\_M\_N\_ani00057\_ani00055\_11149)}&
&\tiny{NRN\_M\_N\_nat00095\_nat00099\_10079)}&
&\tiny{NRD\_S\_N\_sec00011\_sec00016\_00031)}\\
\end{tabular}
\caption{Visual comparison of detected masks. From top to bottom, input pair, manually annotated ground truth masks, predicted masks of using the proposed DMVN method, predicted masks of Alg. \cite{luo2006robust}, \cite{ryu2010detection}; the last two rows are \cite{christlein2012evaluation}'s matched keypoints and predicted masks.}\label{fig.compareCasia}
\end{figure}

\section{Conclusion}\label{sec.conclusions}
\notsure{
In this paper we propose a new deep neural network based solution for the image splicing detection and localization problems. We show that these two problems can be jointly solved using a multitask network in an end-to-end manner, as shown in Fig.~\ref{fig.pipeline}. In particular, we invent the \emph{Deep Dense Matching} layer to find potential splicing regions for two given image features, and we design a \emph{Visual Consistency Validator} module that determines a detection by cross-verifying image content on potential splicing regions. Compared to classic solutions, the proposed approach does not rely on any handcrafted features, heuristic rules and parameters, or extra post-processing, but could fulfill both splicing localization and detection. Our experiments on two very large datasets show that this new approach is much faster and achieves a much higher AUC score than classic approaches, and that it also provides meaningful splicing masks that can help a human conduct further forensics analysis (see Fig.~\ref{fig.visCASIA}). Last but not least, though we train our DMVN model \wrt both localization and detection branches, the proposed DMVN could be trained \wrt only the detection branch while still attaining the capacity to localize splicing masks due to the feed-forward nature of DMVN. This fact means that the proposed DMVN model can be easily finetuned to a new CISD dataset with only label annotations, and that one can save tremendous time and cost for splicing mask annotation in CISD training data collection. }
\newpage
\bibliographystyle{ACM-Reference-Format}
\bibliography{sigproc} 

\end{document}